\documentclass[10pt]{cai}
\usepackage[utf8]{inputenc}
\UseRawInputEncoding

\usepackage[T1]{fontenc}
\usepackage{graphicx}
\usepackage{algorithm}%
\usepackage{algorithmicx}%
\usepackage{algpseudocode}
\usepackage{url}
\usepackage{amsmath}

\begin{document}
\begin{center}


\title{Dynamic Sparse Causal-Attention Temporal Networks for Interpretable Causality Discovery in Multivariate Time Series}

\maketitle

\thispagestyle{empty}

\begin{tabular}{cc}
Meriem Zerkouk \upstairs{\affilone,*}, Miloud Mihoubi\upstairs{\affilone}, Belkacem Chikhaoui\upstairs{\affilone}
\\[0.25ex]
{\small \upstairs{\affilone} Artificial Intelligence Institute, University of TELUQ,  5800, rue Saint-Denis, 
Montreal, Quebec, H2S 3L5, Canada} \\
\end{tabular}
  
\emails{
  \upstairs{*}meriem.zerkouk@teluq.ca 
}
\vspace*{0.2in}

\end{center}

\begin{abstract}
Understanding causal relationships in multivariate time series (MTS) is essential for effective decision-making in fields such as finance and marketing, where complex dependencies and lagged effects challenge conventional analytical approaches. We introduce Dynamic Sparse Causal-Attention Temporal Networks for Interpretable Causality Discovery in MTS (DyCAST-Net), a novel architecture designed to enhance causal discovery by integrating dilated temporal convolutions and dynamic sparse attention mechanisms.
DyCAST-Net effectively captures multiscale temporal dependencies through dilated convolutions while leveraging an adaptive thresholding strategy in its attention mechanism to eliminate spurious connections, ensuring both accuracy and interpretability. A statistical shuffle test validation further strengthens robustness by filtering false positives and improving causal inference reliability.
Extensive evaluations on financial and marketing datasets demonstrate that DyCAST-Net consistently outperforms existing models such as TCDF, GCFormer, and CausalFormer. The model provides a more precise estimation of causal delays and significantly reduces false discoveries, particularly in noisy environments. Moreover, attention heatmaps offer interpretable insights, uncovering hidden causal patterns such as the mediated effects of advertising on consumer behavior and the influence of macroeconomic indicators on financial markets.
Case studies illustrate DyCAST-Net's ability to detect latent mediators and lagged causal factors, making it particularly effective in high-dimensional, dynamic settings. The model's architecture enhanced by RMSNorm stabilization and causal masking ensures scalability and adaptability across diverse application domains.

\end{abstract}

\section{Introduction}
Understanding causal mechanisms in multivariate time series is critical for uncovering the underlying dynamics in complex systems, with far-reaching applications in fields such as healthcare, finance, urban planning, and marketing \cite{wu2024causal}. 

Traditional machine learning models, including decision trees and standard neural networks, are primarily designed to capture correlations rather than directional causal influences. This limitation is significant because correlation, being symmetrical, often fails to reveal the true underlying causes that drive system behavior, potentially leading to misleading conclusions \cite{Chikhaoui2015ANG}.

Historically, controlled interventions have been the gold standard for establishing causality. However, such approaches are often impractical in many real-world scenarios due to ethical, financial, or logistical constraints. Consequently, the rise of observational data combined with sophisticated analytical techniques has spurred research into inferring causal relationships without direct interventions~\cite{Chikhaoui2017DetectingCO}.
Recent advances in deep learning have demonstrated promising directions for causal discovery. Convolutional Neural Networks (CNNs) have been applied to time series data to extract localized features~\cite{nauta2019causal}, while transformer-based models employing self-attention mechanisms have proven effective in capturing long-range dependencies~\cite{vaswani2017attention}. Nevertheless, standard CNNs and vanilla transformers face challenges when dealing with the non-linear, dynamic, and high-dimensional nature of multivariate time series. For instance, conventional CNNs may struggle to model long-term dependencies, and transformers can be computationally demanding and data-hungry.
In this work, we propose DyCAST-Net, a novel hybrid architecture that integrates an enhanced Temporal Convolutional Network (TCN) \cite{Krishna2024AutoIE}  with dynamic sparse \cite{Jiang2024MInference1A}, and multi-head attention mechanisms. Our model leverages dilated convolutions to capture extended temporal dependencies efficiently, while the incorporation of a dynamic sparse attention module allows it to selectively focus on key causal features, reducing noise and redundancy. Additionally, we employ advanced normalization techniques, including RMSNorm and LayerScale, to enhance training stability and improve interpretability.
DyCAST-Net builds upon and extends recent developments in deep learning for causal inference \cite{Mihoubi2024DiscoveringCR} by fusing the localized feature extraction of improved TCNs with the powerful long-range modeling capabilities of transformer-inspired attention. We validate our approach on datasets from the finance and marketing sectors and compare its performance against state-of-the-art methods such as TCDF, CausalFormer, and GCFormer. Experimental results demonstrate that DyCAST-Net not only achieves higher predictive accuracy and computational efficiency but also offers enhanced interpretability through detailed attention score visualizations and delay heatmaps. Our work makes the following significant contributions:
\begin{enumerate}
    \item \textbf{Hybrid TCN--Transformer Architecture with Sparse Attention \cite{Lee2024ATS} and RMSNorm.} 
DyCAST-Net combines a Temporal Convolutional Network (TCN) for short-range features and a Transformer branch with multihead attention 
for long-range dependencies. RMSNorm stabilizes training, while optional attention sparsity boosts efficiency.
    \item \textbf{Localized Feature Cross Modules (LFCM/HFCM) to Merge Multi-scale Context.} 
Custom modules merge multi-scale context by exchanging intermediate features between TCN stages and Transformer blocks via up/down-sampling, blending local (TCN) and global (Transformer) dynamics
    \item \textbf{Enhanced Interpretability through Attention Visualization and Delay Heatmaps.}
Beyond causal predictions, our framework visualizes channel-level attention and temporal delay heatmaps, clarifying learned causal links and time-lagged dependencies.
    \item \textbf{Superior Accuracy and Efficiency on Financial \& Marketing Benchmarks.}
Experiments on financial and marketing datasets show DyCAST-Net surpasses Baselines models in accuracy, efficiency, and fine-grained interpretability.erving interpretability and enabling more fine-grained causal inference.
\end{enumerate}
The remainder of the paper is organized as follows: Section 2 reviews related work in causal discovery and deep learning-based time series analysis. Section 3 details the proposed model and its architecture. Section 4 presents experimental evaluations and comparative analyses, and Section 5 concludes the paper with discussions on future research directions.

\section{Related Work}
Recent advancements in deep learning and attention mechanisms have significantly advanced causal discovery in multivariate time series (MTS), enabling the detection of complex causal patterns and the handling of massive datasets. 

\paragraph{Statistical and graph-based methods.} Methods such as PCMCI \cite{runge2019} integrate conditional independence testing with time series structures to provide an effective multivariate framework, though their scalability is limited by the computational overhead of repeated tests across numerous variables and time steps in high-dimensional settings. 

\paragraph{CNN approaches.}Convolutional neural network approaches, exemplified by the Temporal Causal Discovery Framework (TCDF) \cite{nauta2019causal}, excel at extracting localized temporal features to model short-term dependencies, yet they may struggle with capturing long-range or evolving interactions critical for constructing complete causal graphs.  

\paragraph{Transformer and hybrid models.}Transformer-based architectures \cite{vaswani2017attention} leverage self-attention mechanisms to capture global dependencies by dynamically attending to all time steps and variables, effectively uncovering intricate relationships; however, these models demand extensive computational resources and large datasets to avoid overfitting, posing challenges in resource-constrained scenarios. 
Further advances include recent work leveraging latent contexts from multiple datasets \cite{Gunther2023LatentContexts}, which emphasizes dataset-specific structures, as well as emerging methods such as GCFormer \cite{Xing2023GCFormerGC}, inductive Granger causal modeling \cite{chu2020}, Daring \cite{he2021}, and causal feature selection under latent confounding \cite{Chikhaoui2014PatternbasedCR} that strive to balance scalability, precision, and interpretability while contending with the inherent trade-offs between computational complexity and robust causal inference.

A notable development in this landscape is CausalFormer \cite{Kong2024CausalFormerAI}, an interpretable transformer-based model specifically designed for temporal causal discovery. Unlike earlier work that primarily examines only certain model components (e.g., attention or convolution weights) to infer causality, CausalFormer incorporates a “causality-aware transformer” \cite{Wang2024CausalityAwareTN}   employing multi-kernel causal convolution and a decomposition-based causality detector. By leveraging regression relevance propagation to interpret the trained model’s global structure, CausalFormer constructs detailed causal graphs that capture intricate temporal interactions. Experiments on synthetic, simulated, and real datasets demonstrate its state-of-the-art performance in uncovering temporal causality, underscoring the importance of interpretability in deep learning for causal inference.

Despite these advances, critical challenges persist in temporal causal discovery presented in table \ref{tab:model_comparison}:

\begin{table}[ht]
\centering
\caption{Comparative Analysis of DyCAST-Net, TCDF, CausalFormer, and GCformer}
\label{tab:model_comparison}
\begin{tabular}{|p{2.1cm}|p{3.1cm}|p{3cm}|p{3.1cm}|p{3.1cm}|}

\textbf{Feature} & \textbf{DyCAST-Net} & \textbf{TCDF} & \textbf{CausalFormer} & \textbf{GCformer} \\ \hline

\textbf{Architecture} & 
\begin{tabular}[c]{@{}l@{}}Dilated Convolutions\\ + Multi-head Attention\\ + Dynamic Sparsity\end{tabular} &
\begin{tabular}[c]{@{}l@{}}Dilated Convolutions\\ + Attention Mechanism\end{tabular} &
\begin{tabular}[c]{@{}l@{}}Transformer\\ + Self-Attention\end{tabular} &
\begin{tabular}[c]{@{}l@{}}GCN\\ + Transformer\end{tabular} \\ \hline

\textbf{Attention Type} & 
\begin{tabular}[c]{@{}l@{}}Multi-head with\\ Dynamic Sparsity\end{tabular} &
Local Attention & 
Global Self-Attention & 
Graph + Self-Attention \\ \hline

\textbf{Normalization} & 
\begin{tabular}[c]{@{}l@{}}RMSNorm\\ + LayerScale\end{tabular} & 
LayerNorm & 
LayerNorm & 
BatchNorm \\ \hline

\textbf{Key Components} &
\begin{tabular}[c]{@{}l@{}}- Depth-wise Convolutions\\ - Dynamic Pruning\\ - Skip Connections\\ - Causal Masking\end{tabular} &
\begin{tabular}[c]{@{}l@{}}- Dilated Convolutions\\ - Attention for Delay\\ \hspace{0.25cm} Estimation\end{tabular} &
\begin{tabular}[c]{@{}l@{}}- Multi-head Attention\\ - Positional Encoding\end{tabular} &
\begin{tabular}[c]{@{}l@{}}- Graph Convolution\\ - Temporal Transformer\end{tabular} \\ \hline

\textbf{Handling Long Dependencies} & 
\begin{tabular}[c]{@{}l@{}}Dilated Convolutions\\ + Sparse Attention\end{tabular} & 
Dilated Convolutions & 
Self-Attention & 
GCN + Dilated Convolutions \\ \hline

\textbf{Dynamic Sparsity} & 
Yes (Threshold-based) & 
No & 
No & 
No \\ \hline

\textbf{Computational Efficiency} & 
High (Sparse Attention) & 
Medium & 
Low (Dense Attention) & 
Medium \\ \hline

\textbf{Interpretability} & 
\begin{tabular}[c]{@{}l@{}}Attention Scores\\ + Delay Heatmaps\end{tabular} & 
Attention Weights & 
Attention Maps & 
Graph Edges \\ \hline

\textbf{Strengths} & 
\begin{tabular}[c]{@{}l@{}}- Efficient Sparse-Attention\\ - Stable Training\\ - Explicit Delay\\ \hspace{0.25cm} Estimation\end{tabular} & 
\begin{tabular}[c]{@{}l@{}}- Temporal Delay\\ \hspace{0.25cm} Detection\\ - Simple Architecture\end{tabular} & 
\begin{tabular}[c]{@{}l@{}}- Long-Range\\ \hspace{0.25cm} Dependencies\\ - Non-Linear Effects\end{tabular} & 
\begin{tabular}[c]{@{}l@{}}- Spatial-Temporal\\ \hspace{0.25cm} Modeling\\ - Graph-Based Causality\end{tabular} \\ \hline

\textbf{Limitations} & 
\begin{tabular}[c]{@{}l@{}}- Tuning Sparse\\ \hspace{0.25cm} Thresholds\end{tabular} & 
\begin{tabular}[c]{@{}l@{}}- Limited to Local\\ \hspace{0.25cm} Contexts\end{tabular} & 
\begin{tabular}[c]{@{}l@{}}- Computationally Heavy\\ - Data-Hungry\end{tabular} & 
\begin{tabular}[c]{@{}l@{}}- Requires Predefined\\ \hspace{0.25cm} Graph Structure\end{tabular} \\ 
\end{tabular}
\end{table}

\begin{itemize}
\item \textbf{Multi-scale Dependency Modeling}: CNN-based approaches like TCDF \cite{nauta2019causal} excel at local features but neglect global interactions, while Transformers \cite{vaswani2017attention} capture long-range dependencies at the cost of local granularity.
\item \textbf{Dynamic Sparsity}: Most attention mechanisms \cite{Xing2023GCFormerGC} employ dense computations, retaining noisy connections in high-dimensional MTS \cite{Kong2024CausalFormerAI}.
\item \textbf{Interpretability-Complexity Trade-off}: Though methods like CausalFormer \cite{Kong2024CausalFormerAI} improve transparency, their reliance on dense attention obscures lag-specific causal pathways.
\item \textbf{Resource Constraints}: Graph-based methods (e.g., PCMCI \cite{runge2019}) scale quadratically with variables, while Transformers require extensive data to avoid overfitting \cite{he2021}.
\end{itemize}

\section{Causal Discovery paradigm Using DyCAST-Net}
The DyCAST-Net is designed to uncover causal relationships in multivariate time series (MTS) data. The paradigm leverages a combination of multi-head attention mechanisms and convolutional neural networks to process temporal data and construct a Temporal Causal Graph (TCG). This graph represents the discovered causal relationships and their associated delays, providing insights into the temporal dynamics of the system under study.
Figure 1 illustrates the architecture of DyCAST-Net, which operates using $N$ independent parallel networks. Each network has the same structure but processes different input variables. 

\subsection{Problem Statement}
The task of temporal causal discovery from MTS data involves analyzing a dataset $X$ composed of $N$ continuous time series, each with a uniform length $T$ (i.e., $X = \{X_1, X_2, \dots, X_N\} \in \mathbb{R}^{N \times T}$).

The primary goal is to uncover the causal relationships that exist among these time series, quantify the time delays between causes and their corresponding effects, and represent these findings within a temporal causal graph.
In this approach, a directed causal graph $G = (V, E)$ is constructed, where each vertex $v_i \in V$ represents a distinct time series $X_i$. A directed edge $e_{i,j} \in E$ from vertex $v_i$ to vertex $v_j$ indicates that $X_i$ causally influences $X_j$. Additionally, the graph may include paths $p = \{v_i, \dots, v_j\}$, which denote the sequence of causal influence from $v_i$ to $v_j$. Importantly, each edge $e_{i,j}$ in the graph is labeled with a weight $d(e_{i,j})$, representing the time delay between the initial cause $X_i$ and the resulting effect on $X_j$.


\subsection{Our Proposed Architecture (DyCAST-Net)}
\label{subsec:our_architecture}

\begin{figure}[H]
\centering
\includegraphics[width=0.81\linewidth]{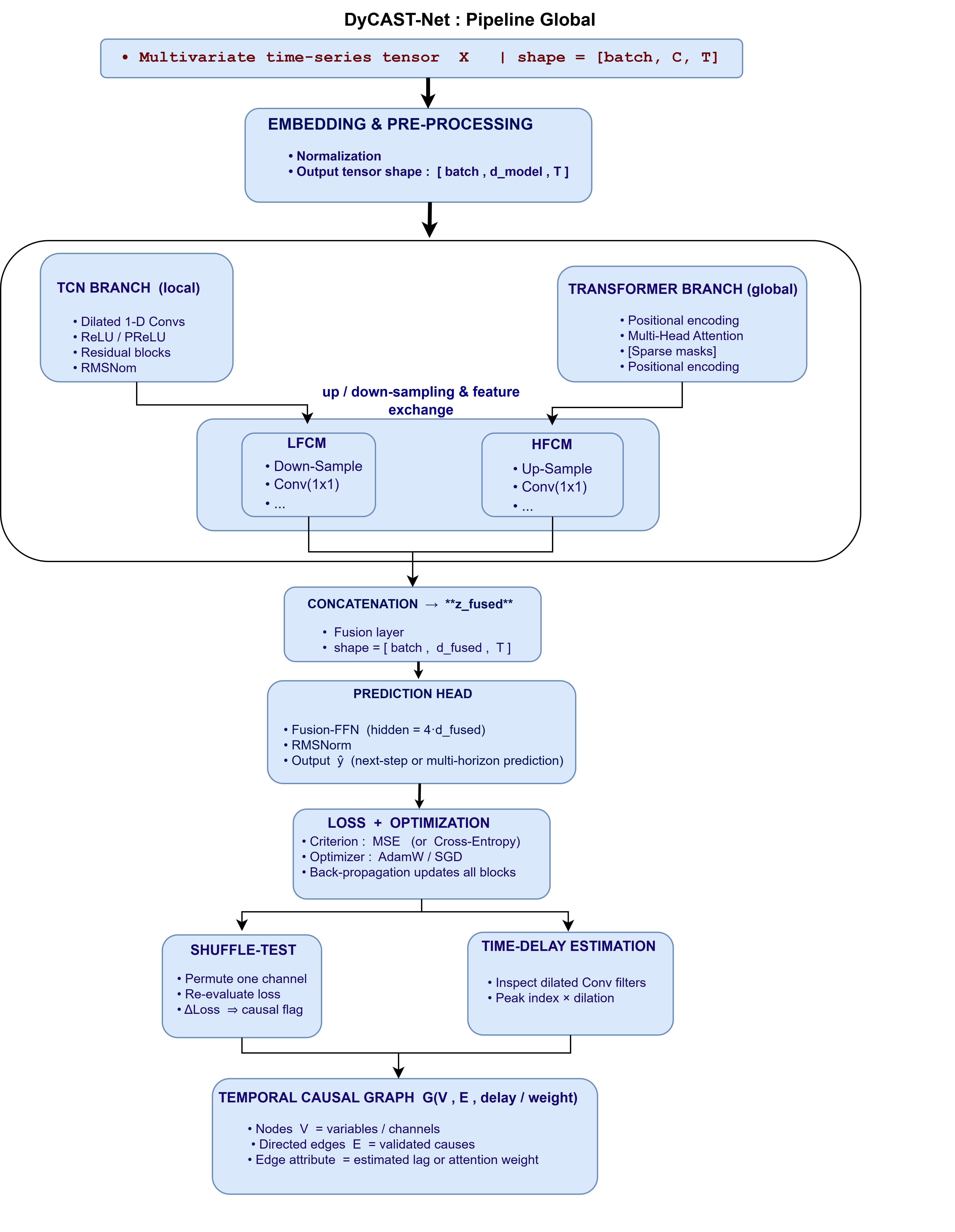}
\caption{Illustration of DyCAST-Net's training and optimization process.}
\label{fig:training-optimization-process}
\end{figure}

\noindent
We propose \textbf{DyCAST-Net}, a novel deep learning architecture designed to uncover causal relationships in \emph{multivariate time series} (MTS)  
Our approach starts from raw time series data (with $N$ variables/channels and $T$ time steps each) and culminates in a \emph{causal graph} whose nodes represent the original variables, while directed edges denote discovered causal influences (along with an estimated time delay or strength). The architecture integrates:

\begin{itemize}
  \item \textbf{Dilated Convolution Blocks} (a TCN-style backbone) to capture long-range temporal dependencies.
  \item \textbf{Multi-Head Attention} to highlight the most important temporal elements (\emph{positions}) and channels (\emph{variables})~\cite{vaswani2017attention}.
  \item \textbf{Dynamic Sparse Attention}, which prunes small attention weights to focus on dominant causal signals.
  \item \textbf{RMSNorm}
  or \textbf{LayerNorm}
  for stable training, optionally combined with \emph{LayerScale} for better gradient flow.
  \item \textbf{Skip Connections} to preserve original signals and improve gradient propagation.
\end{itemize}
Figure \ref{fig:training-optimization-process} illustrates this pipeline, from raw MTS input to causal graph construction.

\subsubsection{Dilated Convolution Blocks with Skip Connections}
We employ an \emph{N}-channel \textit{Temporal Convolutional Network} (TCN)
, stacking several layers (\emph{DyCAST-Block}) in sequence. Each block contains:
\begin{enumerate}
    \item \textbf{Depthwise Dilated Convolution} (stride=1) with a chosen kernel size (e.g., 2) and dilation factor $d=2^l$ at layer $l$, to expand the receptive field.
    \item \textbf{Causal Masking} (a ``chomp'' operation) ensuring the convolution only uses past and present information, preserving temporal causality.
    \item \textbf{Normalization} via RMSNorm or LayerNorm. RMSNorm
    normalizes based on the RMS (root mean square) of each channel, while LayerNorm
    normalizes across features. One can optionally add \emph{LayerScale} if desired.
       \item \textbf{Residual / Skip Connection} with a non-linear activation, enabling deeper architectures without vanishing gradients.
\end{enumerate}
Stacking $L$ such blocks allows DyCAST-Net to model long-range dependencies effectively.
\subsubsection{Multi-Head Attention with Dynamic Sparsity}
\label{subsubsec:attention_sparsity}
In each \emph{DyCAST-Block}, we integrate a \textbf{multi-head attention} module~\cite{vaswani2017attention}:
\[
\mathrm{Attn}(\mathbf{Q}, \mathbf{K}, \mathbf{V}) 
= \mathrm{softmax}\Big(\frac{\mathbf{Q}\mathbf{K}^\top}{\sqrt{d_{\mathrm{head}}}} + \mathrm{Mask}\Big)\,\mathbf{V},
\]
where \(\mathbf{Q}, \mathbf{K}, \mathbf{V}\) are projected embeddings of the block’s features. A \emph{causal mask} ensures no attention to future time steps. 
Additionally, to promote interpretability and reduce noise, we use \textbf{Dynamic Sparse Attention}: any attention weight below a threshold is zeroed out (\emph{pruning}) so that the model focuses on the most significant relationships. This pruning is especially useful in highlighting potential causal links.
\subsubsection{Channel-Level Weighting for Variable Selection}
Before or after the convolution blocks, a learnable \emph{channel-level attention vector} $\boldsymbol{\alpha}\in\mathbb{R}^{N}$ (one weight per channel) modulates the input data:
\[
\mathbf{X}_{\mathrm{weighted}} = \mathbf{X} \odot \mathrm{Softmax}(\boldsymbol{\alpha}),
\]
where $\mathbf{X}$ is the original MTS, and $\mathrm{Softmax}(\boldsymbol{\alpha})$ ensures the weights sum to 1 and remain positive. Channels whose corresponding learned weights are large are more influential in predicting the target.

\subsubsection{Training and Objective}
DyCAST-Net is trained via a standard \textbf{mean squared error} (MSE) loss for next-step (or multi-step) prediction. We may add $\ell_1$ regularization to (a) the channel-level weights $\boldsymbol{\alpha}$ or (b) convolution kernels and attention matrices, thus encouraging \emph{channel sparsity} or \emph{attention sparsity}, respectively:
\[
\mathcal{L}(\Theta) 
= \frac{1}{T_\mathrm{eff}} \sum_{t=1}^{T_\mathrm{eff}} \bigl(\hat{y}_t - y_t\bigr)^2 
+ \lambda_K \|\text{Kernels}\|_1 
+ \lambda_M \|\text{Masks}\|_1,
\]
where $T_\mathrm{eff}$ is the effective prediction range, and $(\hat{y}_t, y_t)$ are predicted and true target values at time $t$. Minimizing this loss encourages an accurate yet parsimonious (i.e., sparser) model.

\subsection{Causality Detection}
\label{subsec:causality_detection}

Once the DyCAST-Net model is trained to predict a specific target variable, we analyze the learned parameters to:
\begin{enumerate}
    \item \textbf{Identify causal channels} (which input variables most strongly affect the target),
    \item \textbf{Estimate causal delays} (how far in the past each cause matters),
    \item \textbf{Construct a directed causal graph} that encodes these relationships.
\end{enumerate}

\subsubsection{Channel Selection and Shuffle Test}
To extract \emph{potential causes}, we first examine the \emph{channel-level attention vector} $\boldsymbol{\alpha}$ (Sec.~\ref{subsubsec:attention_sparsity}). Any channel $j$ whose weight $\alpha_j$ exceeds a threshold $\tau$ is considered a \emph{candidate cause}. Next, to validate these candidates, we use a \emph{shuffle test}:
\begin{itemize}
    \item Randomly permute the time points of channel $j$.
    \item Recompute the prediction error $\mathcal{L}_{\mathrm{test}}$ with this permuted input.
    \item If $\mathcal{L}_{\mathrm{test}}$ degrades substantially compared to the baseline, channel $j$ likely has a genuine causal influence on the target.
\end{itemize}

\subsubsection{Delay Estimation from Dilated Convolution Kernels}
A key advantage of TCN-based models is that \emph{dilated convolution filters} directly reveal how far back in time a variable contributes. For each validated channel $j$, we examine the learned convolution filter parameters $\mathbf{w}_{j}$ at each block/layer. The \textit{index of the maximum absolute weight} indicates the dominant receptive field location. Combining that with the dilation factor $d$ yields the approximate \emph{time delay}:
\[
\mathrm{delay}_{j\to \text{target}} \approx \mathrm{arg\,max}_{k} \big|\mathbf{w}_{j}[k]\big|\; \times d.
\]
This can be aggregated if multiple layers contribute.

\subsubsection{Constructing the Causal Graph}
\label{subsubsec:graph_construction}
To form the \textbf{causal graph}, we do the following for \emph{each target} variable in the MTS:
\begin{itemize}
    \item \textbf{Node Definition}: each channel/variable $v \in \{1,2,\dots,N\}$ is a node in the graph.
    \item \textbf{Edge Definition}: if a channel $j$ is found to be a cause of the target $i$, we add a directed edge $(j \rightarrow i)$.
    \item \textbf{Edge Weights (Delays)}: each edge is labeled with the discovered time delay or an attention-based importance score. Thus,
    \[
    (j \to i) \quad \text{has edge attribute (delay) } = \mathrm{delay}_{j \to i}.
    \]
\end{itemize}
The final result is a directed graph $G = (V, E)$ that encodes the \emph{who-causes-whom} relationships along with \emph{how far in the past} they exert influence. Typically, large delays or minimal attention scores might be pruned to yield a more concise causal subgraph.

\bigskip

\noindent\textbf{Remark on Interpretation.} 
In this causal graph:
\begin{itemize}
    \item \textbf{Nodes} represent individual time series (variables) in the dataset.
    \item \textbf{Directed edges} indicate the discovered causal effect from one variable to another.
    \item \textbf{Edge values (labels)} represent the \emph{estimated time lag} (delay), or alternatively the \emph{confidence/strength} derived from final attention weights.
\end{itemize}
Hence, DyCAST-Net provides an \emph{end-to-end} pipeline: from raw multivariate time series to a concise, interpretable directed graph detailing the likely causes, their strengths, and their temporal offsets.
\section{Experiments Result and Discussion}
To rigorously evaluate DyCAST-Net’s capabilities in temporal causal discovery, we conduct extensive experiments addressing three critical questions:
Accuracy: Does DyCAST-Net outperform state-of-the-art baselines in identifying causal relationships and estimating time delays?
Robustness: How does the model handle noise and high-dimensional regimes?
Interpretability: Do attention patterns align with domain-specific causal mechanisms?
We evaluate DyCAST-Net on financial time series (Fama-French Three-Factor Model) and fMRI data, comparing against table \ref{tab:model_comparison}. Metrics include F1-score, precision, recall, and Delay Estimation Accuracy (DEA). 
  
\subsection{Dataset}
We validate our model on two benchmarks: Finance and FMRI. The Finance benchmark includes 10 datasets generated using the Fama-French Three-Factor Model \cite{Kleinberg2013}, focusing on factors like volatility, size, and value. We excluded one dataset lacking causal relationships. The FMRI benchmark features 28 simulated BOLD datasets \cite{Smith2011}, each representing distinct brain networks and tracking neural activity across various regions. We excluded one dataset with 50 nodes due to processing limitations. These selections ensure that our model is tested on relevant and manageable datasets.
The preprocessing of data for DyCAST-Net involves several crucial steps. For the Finance dataset, we normalized the time series using the z-score method. For the FMRI dataset, we applied a low-pass filter to reduce noise. The data were then converted into PyTorch tensors of dimension $\mathbf{X} \in \mathbb{R}^{N \times T}$, where $N$ is the number of time series and $T$ is the number of time steps, in accordance with the Input Representation and Preprocessing stage of the DyCAST-Net model.

\subsection{Experimental Setup}
\paragraph{Hardware Configuration:}
All experiments were performed on a machine with the following specifications:
\begin{itemize}
    \item \textbf{Processor:} Intel Core i9-10700K @ 3.60GHz (12 cores, 24 threads)
    \item \textbf{Memory:} 64GB DDR4 RAM @ 3200MHz
    \item \textbf{Graphics Card:} NVIDIA GeForce RTX 3090 with 24GB GDDR6X
    \item \textbf{Storage:} 1TB NVMe SSD and \textbf{Operating System:} Windows 11
\end{itemize}
\paragraph{Evaluation protocol.}
We use 5-fold \emph{expanding-window} cross-validation: fold $i$ trains on
the first $60+5i$\,\% of the timeline and tests on the next 10 \%.
Hyper-parameters (kernel, dilation base, learning rate) are optimised on
fold 0 via Bayesian search (50 trials). Early stopping (patience 15)
relies on validation loss. 
are retrained on the \emph{same} splits with seed 1111.

The configuration of key hyperparameters for training the  DyCAST-Net are presented in table \ref{tab:training_config},  model was designed to ensure optimal learning while capturing complex temporal dependencies. 
\begin{table}[h]
\centering
\caption{Training Configuration Parameters}
\label{tab:training_config}
\begin{tabular}{|p{3cm}|p{9.5cm}|}

\textbf{Parameter} & \textbf{Value} \\ \hline
CUDA Enabled & True (training conducted with GPU acceleration) \\ \hline
Number of Epochs & 2000 (for prolonged and comprehensive learning) \\ \hline
Kernel Size & 6 (optimized for capturing diverse temporal dependencies) \\ \hline
Number of Levels & 2 (enhancing the modeling of complex relationships) \\ \hline
Learning Rate & 0.001 (using the Adam optimizer for fine-tuned updates) \\ \hline
Dilation Rate & 4 (expanding the receptive field for long-range dependency modeling) \\ \hline
Log Interval & 500 (for detailed monitoring of model performance) \\ \hline
Seed & 1111 (ensuring reproducibility of results) \\ \hline
Significance Level & 0.5 (threshold for determining causal relationships) \\ \hline
\end{tabular}
\end{table}

\subsection{Layer-wise specification}

Table~\ref{tab:archi} specifies each layer of DyCAST-Net.
The number of heads $H$ is automatically set to the number of series $N$ (so that
\texttt{embed\_dim}=$N$ remains divisible by $H$).

\begin{table}[ht]
\centering
\caption{Layer-by-layer specification of DyCAST-Net
($N$ = number of time-series / channels).}
\label{tab:archi}
\begin{tabular}{ccccccccc}
\toprule
\textbf{Block} & \textbf{Type} & $k$ & \textbf{Dilation} & \textbf{Heads $H$} &
\textbf{Norm} & \textbf{LayerScale} & $\tau_{\text{sparse}}$ & \textbf{Params} \\
\midrule
0 & Depthwise Conv1d & 4 & $2^{0}$ & $N$ & LayerNorm & --              & 0.01 & $4N$ \\
1 & Depthwise Conv1d & 4 & $2^{1}$ & $N$ & LayerNorm & --              & 0.01 & $4N$ \\
2 & Depthwise Conv1d & 4 & $2^{2}$ & $N$ & RMSNorm   & $\gamma = 10^{-4}$ & 0.01 & $4N$ \\
\bottomrule
\multicolumn{9}{l}{\footnotesize
$k$: kernel size, $\tau_{\text{sparse}}$: dynamic pruning threshold (weights < $\tau$ are set to 0).}
\end{tabular}
\end{table}

\paragraph{Experimental protocol.}
For every dataset we perform \textbf{5-fold expanding-window cross-validation}:
fold~$i$ trains on the first $60+5i$\,\% of the timeline and tests on the next
10 \%.  Hyper-parameters ($k$, dilation base, learning rate) are optimised on
fold 0 via Bayesian search (50 trials).  Early stopping is triggered if the
validation loss does not improve for 15 epochs.  All models—including TCDF,
CausalFormer and GCFormer—are re-trained with \textbf{the exact same splits}.

\subsection{Causal Discovery Results}
The following section provides a detailed result of the causal discovery results,
we examine the inferred causal graphs at multiple training epochs and for different target nodes.
Figures 3–5 illustrate how certain causal links evolve over training, 
The following section provides a detailed result of the causal discovery results obtained from our model on the Finance and fMRI datasets. The analysis is based on the attention heatmaps and causal graphs generated at different training epochs (100, 300, and the complete graph).
\begin{figure}[ht!]
\centering
\begin{subfigure}[b]{0.32\textwidth}
    \includegraphics[width=\textwidth]{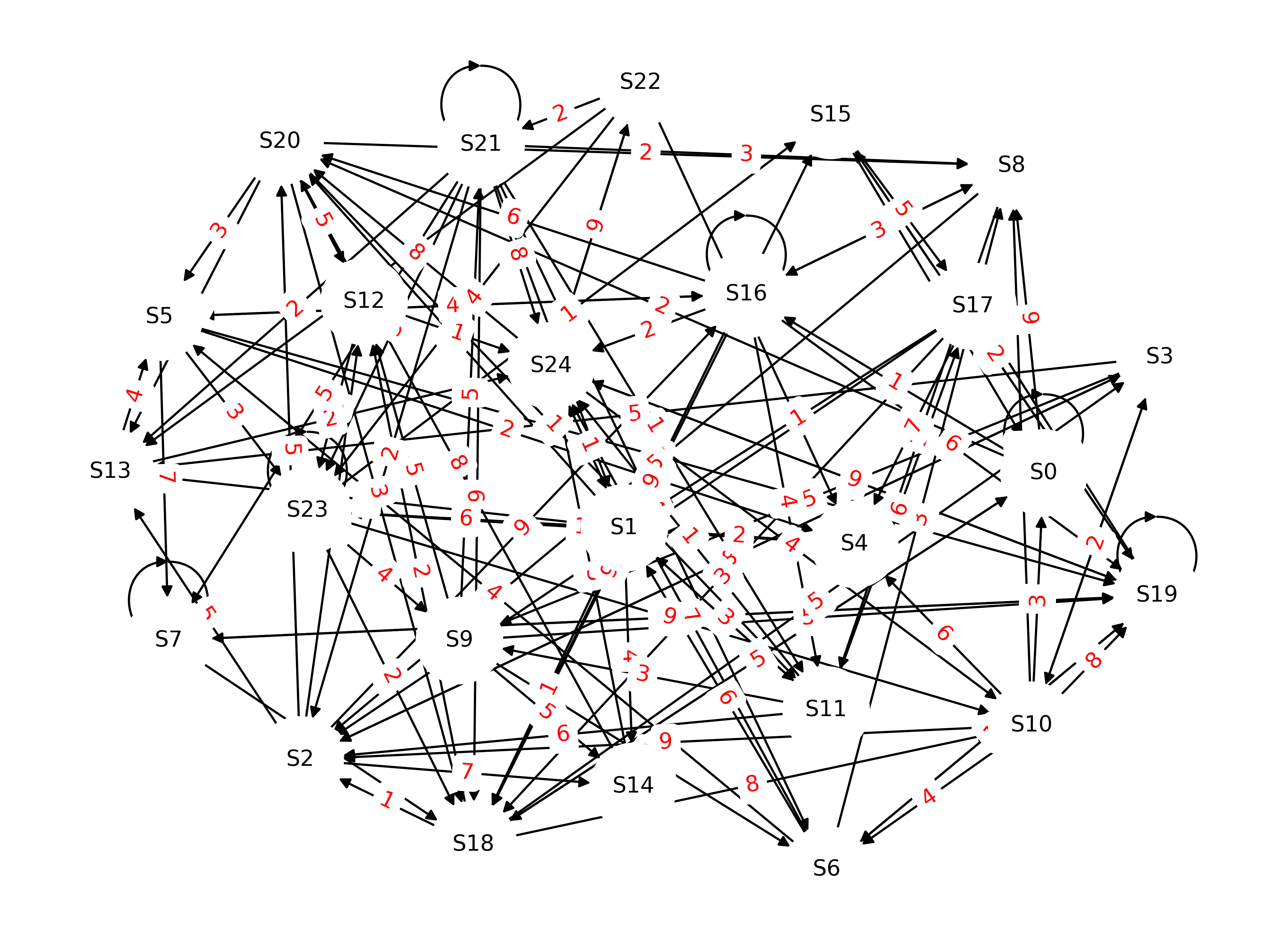}
\end{subfigure}
\hfill
\begin{subfigure}[b]{0.32\textwidth}
    \includegraphics[width=\textwidth]{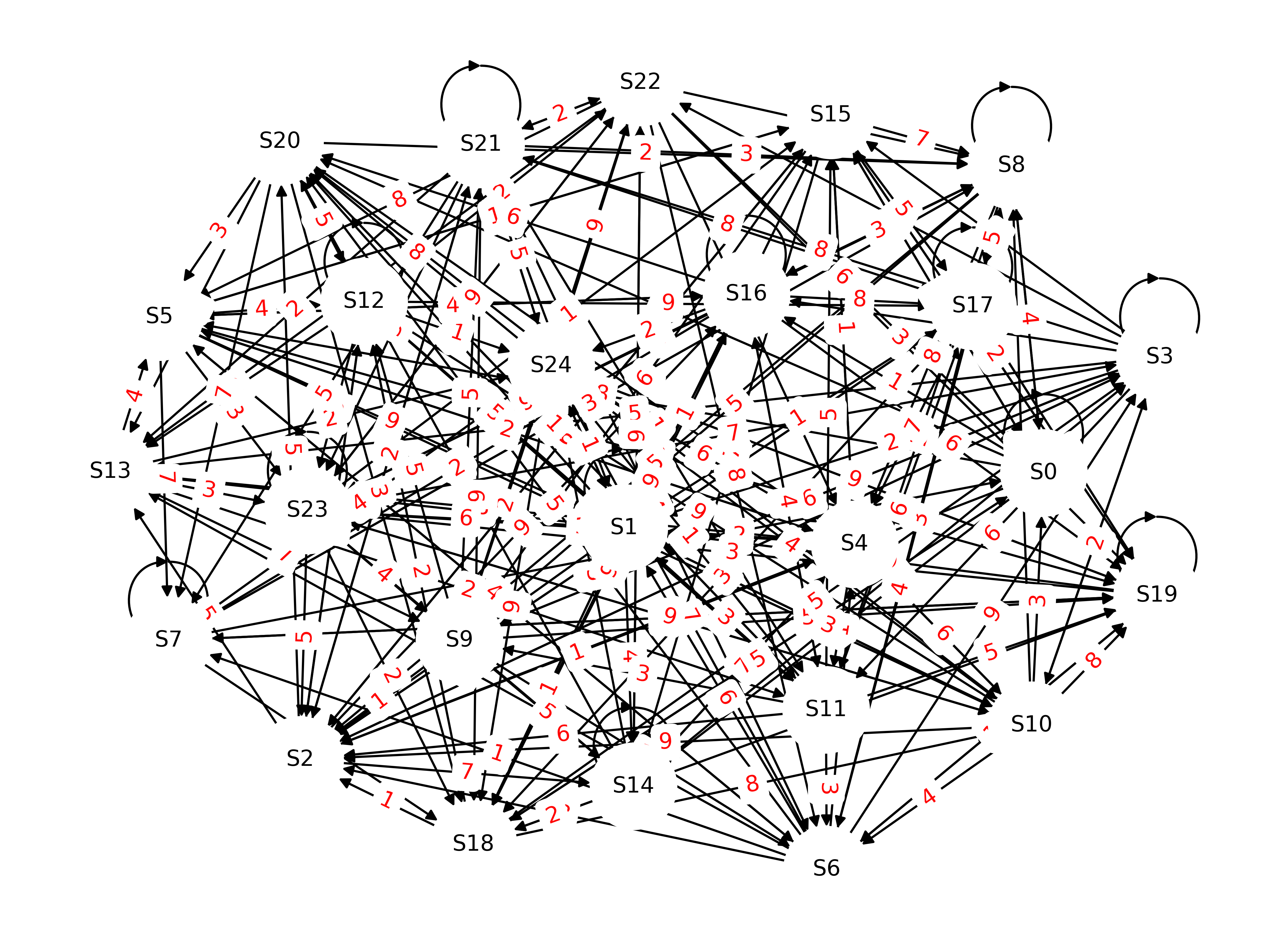}
\end{subfigure}
\hfill
\begin{subfigure}[b]{0.32\textwidth}

    \includegraphics[width=\textwidth]{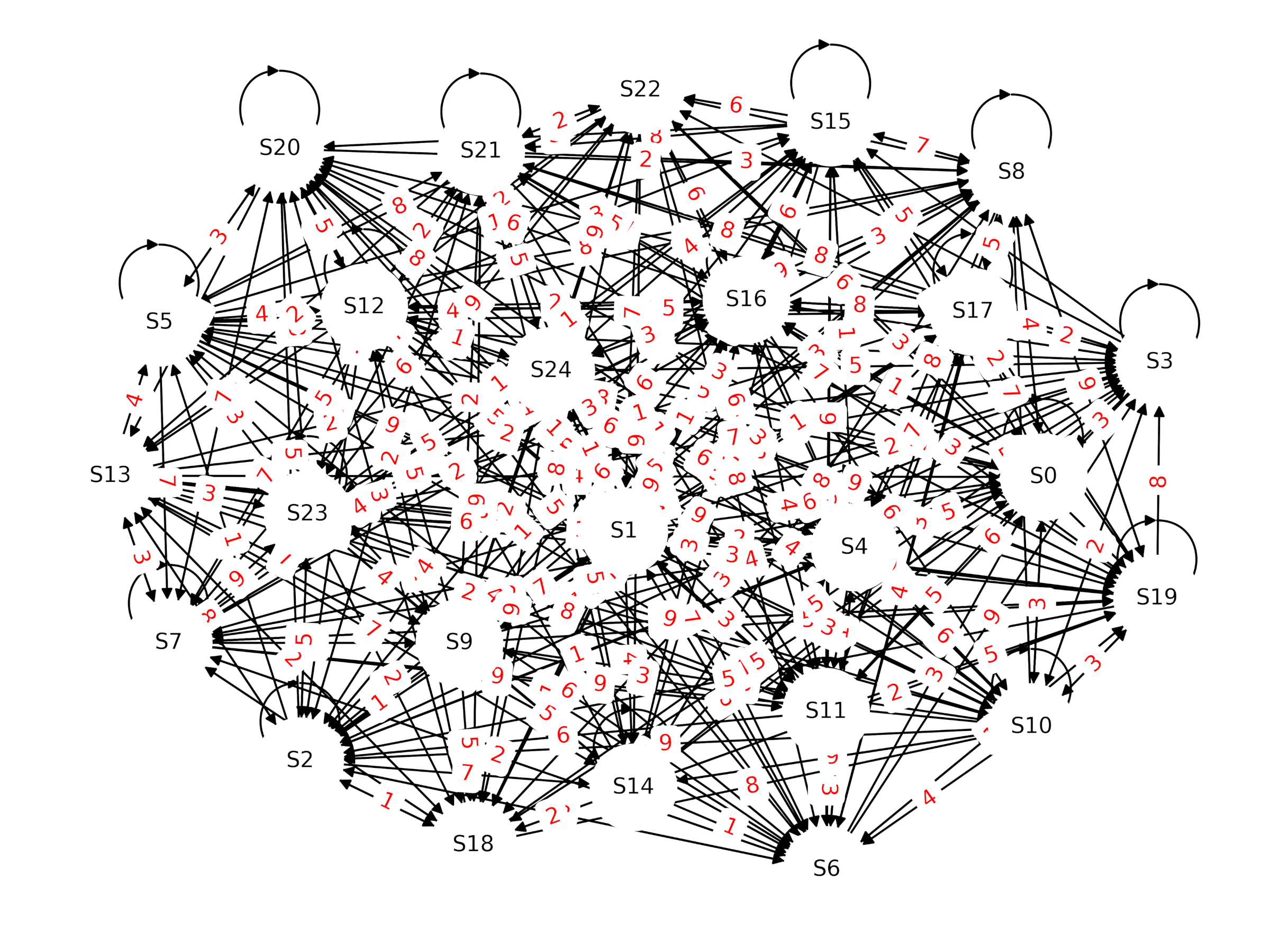}
\end{subfigure}
\caption{Causal relationships in finance data over different epochs (Left: 100, Middle: 300, Right: 2000)}
\label{fig:ligne1}
\end{figure}

The Finance dataset consists of 25 time series, each representing a financial variable. 
The causal graphs at different epochs (Figures 3 depict the progressive discovery of causal relationships. At 100 epochs, the graph is relatively sparse, indicating that only the most dominant causal relationships are detected. By 300 epochs, the graph becomes denser, revealing additional causal interactions. The final graph presents a comprehensive structure, capturing the full extent of causal dependencies in the dataset.
The attention heatmap (Figure 5 ( left) illustrates the intensity of causal influence among the time series. The color distribution in the heatmap highlights strong dependencies, with yellow indicating high attention scores and dark purple representing weak or no relationships.

\begin{figure}[ht!]
\centering
\begin{subfigure}[b]{0.32\textwidth}
    \includegraphics[width=\textwidth]{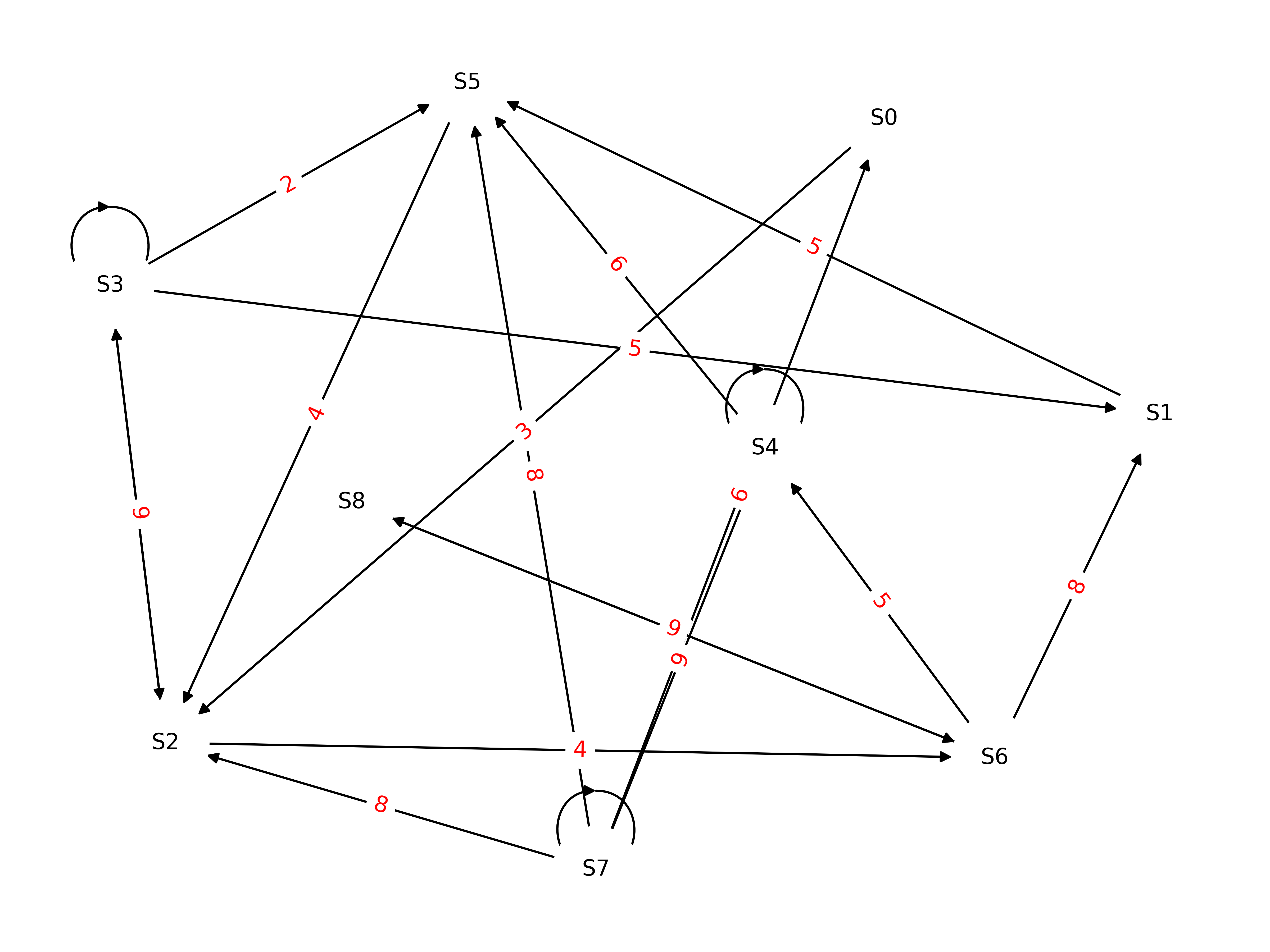}
\end{subfigure}
\hfill
\begin{subfigure}[b]{0.32\textwidth}
    \includegraphics[width=\textwidth]{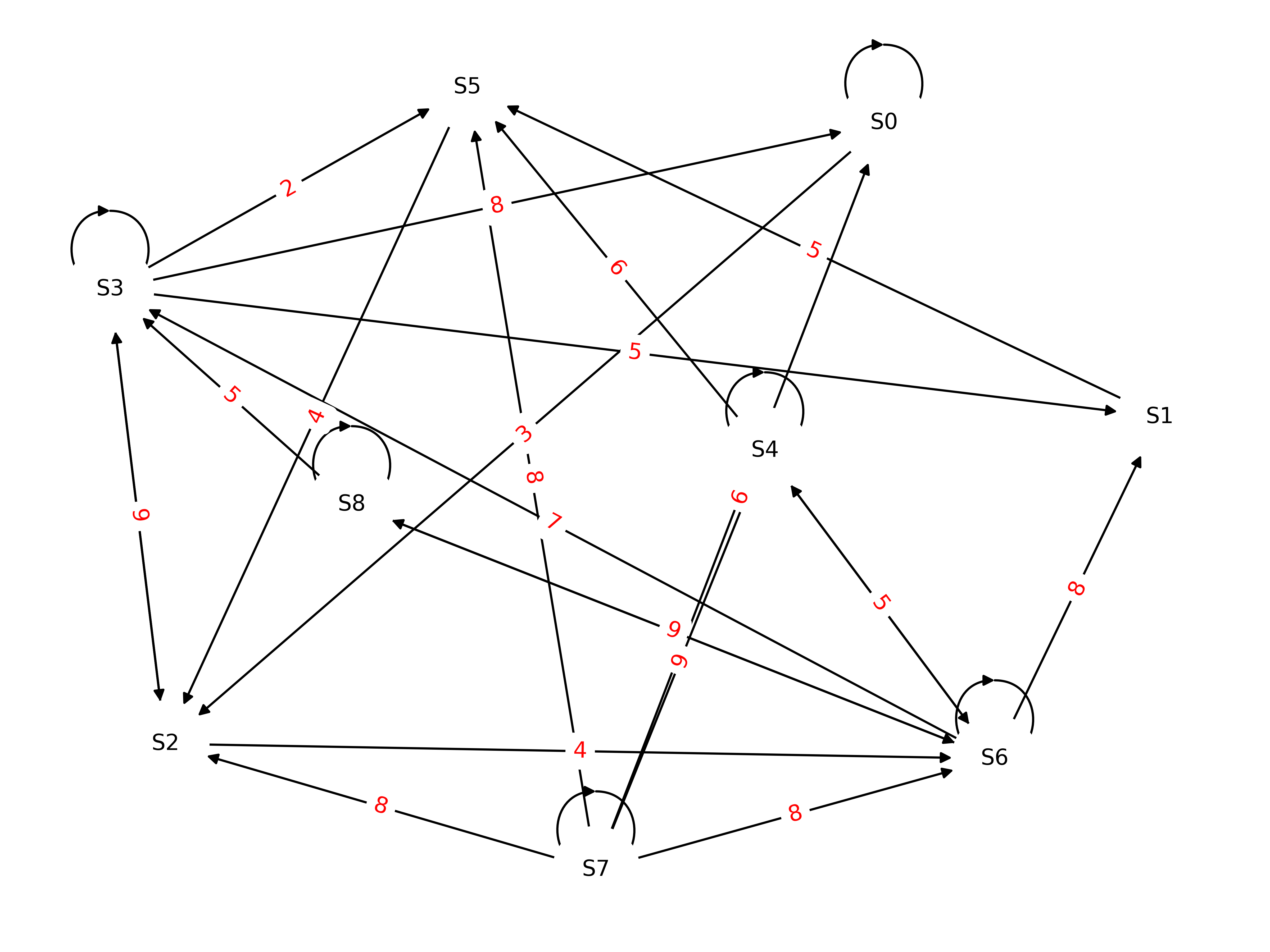}
\end{subfigure}
\hfill
\begin{subfigure}[b]{0.32\textwidth}

    \includegraphics[width=\textwidth]{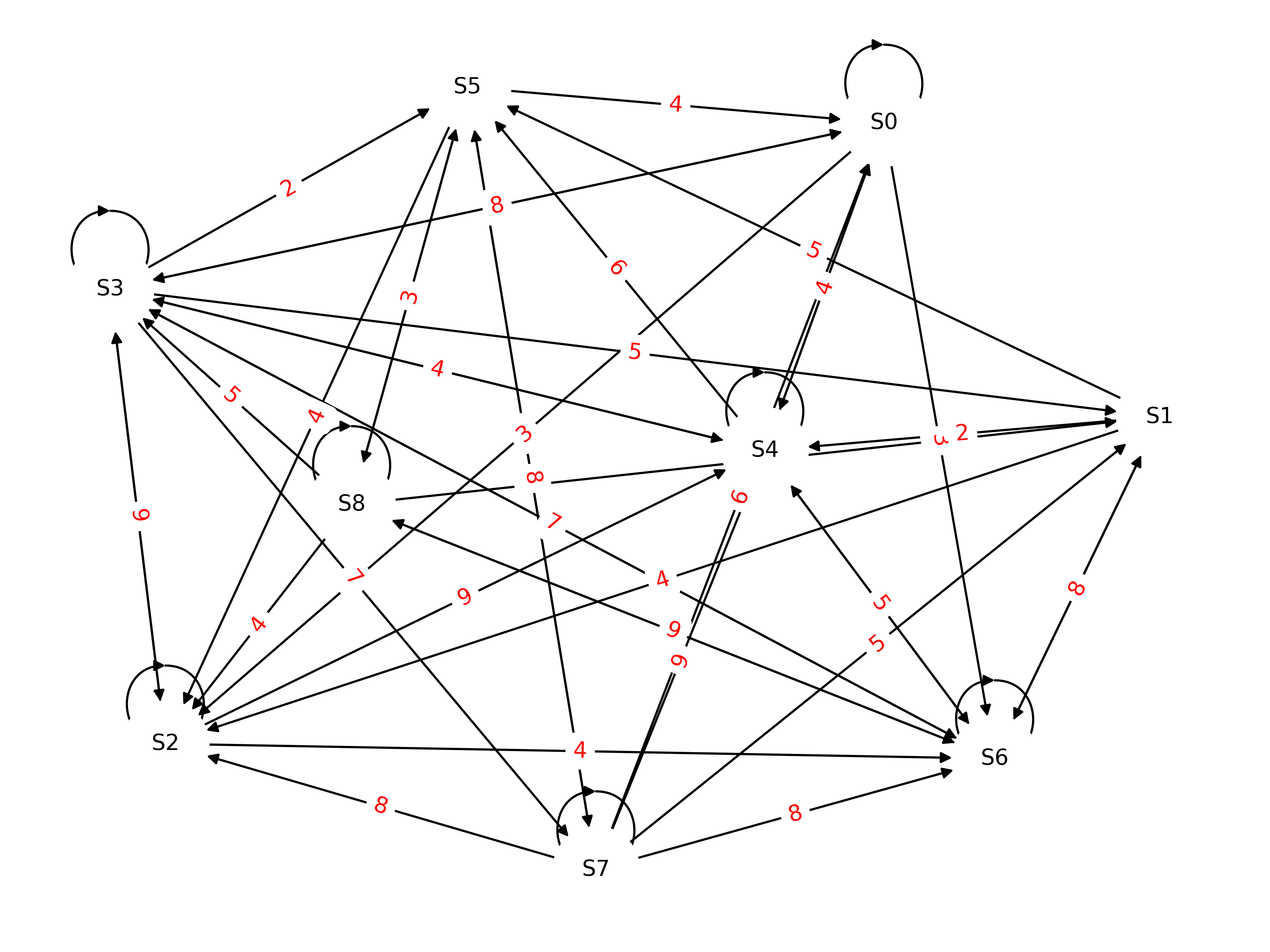}
\end{subfigure}
\caption{Causal relationships in RMFI data over different epochs (Left: 100, Middle: 300, Right: 2000)}
\label{fig:ligne2}
\end{figure}

\subsection{Analysis of the fMRI Dataset}

Similar to the Finance dataset, the causal graphs for the fMRI dataset (Figures 4)  show the evolution of causal discovery over training epochs. At 100 epochs, only a few dominant connections are identified. By 300 epochs, the graph complexity increases as more relationships are discovered. The complete graph encapsulates the full causal structure, offering a comprehensive representation of brain connectivity.
The fMRI dataset also consists of sample of 9 time series, representing neural activity measurements. The attention heatmap (Figure  reveals patterns of interactions between different brain regions. Higher attention scores (yellow regions) suggest strong causal influence, which aligns with expected neural connectivity patterns.

\begin{figure}[ht!]
\centering
\begin{subfigure}[b]{0.45\textwidth}
    \includegraphics[width=\textwidth]{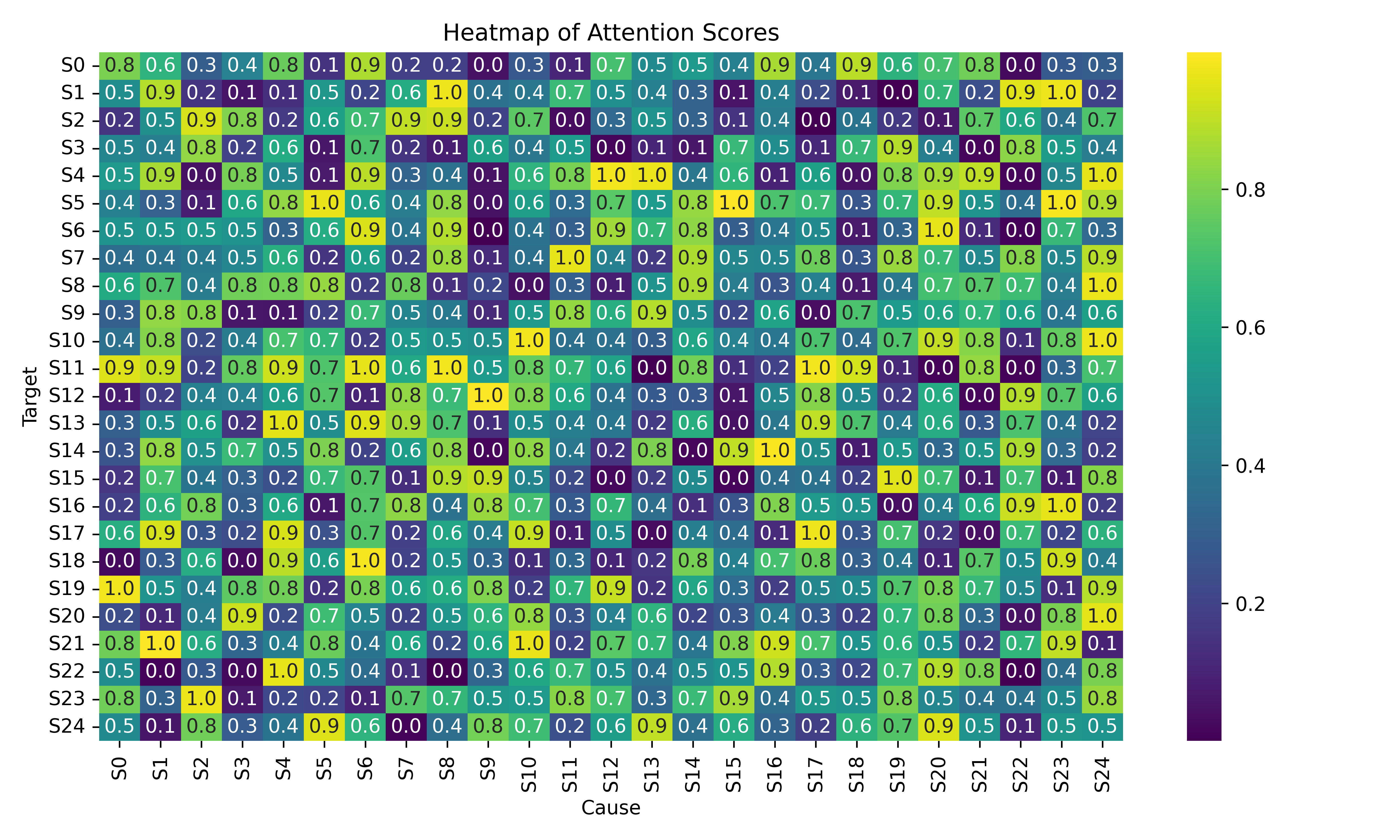}
    \caption{Finance-25 variables.}
    \label{fig:finance_heatmap}
\end{subfigure}
\hfill
\begin{subfigure}[b]{0.45\textwidth}
    \includegraphics[width=\textwidth]{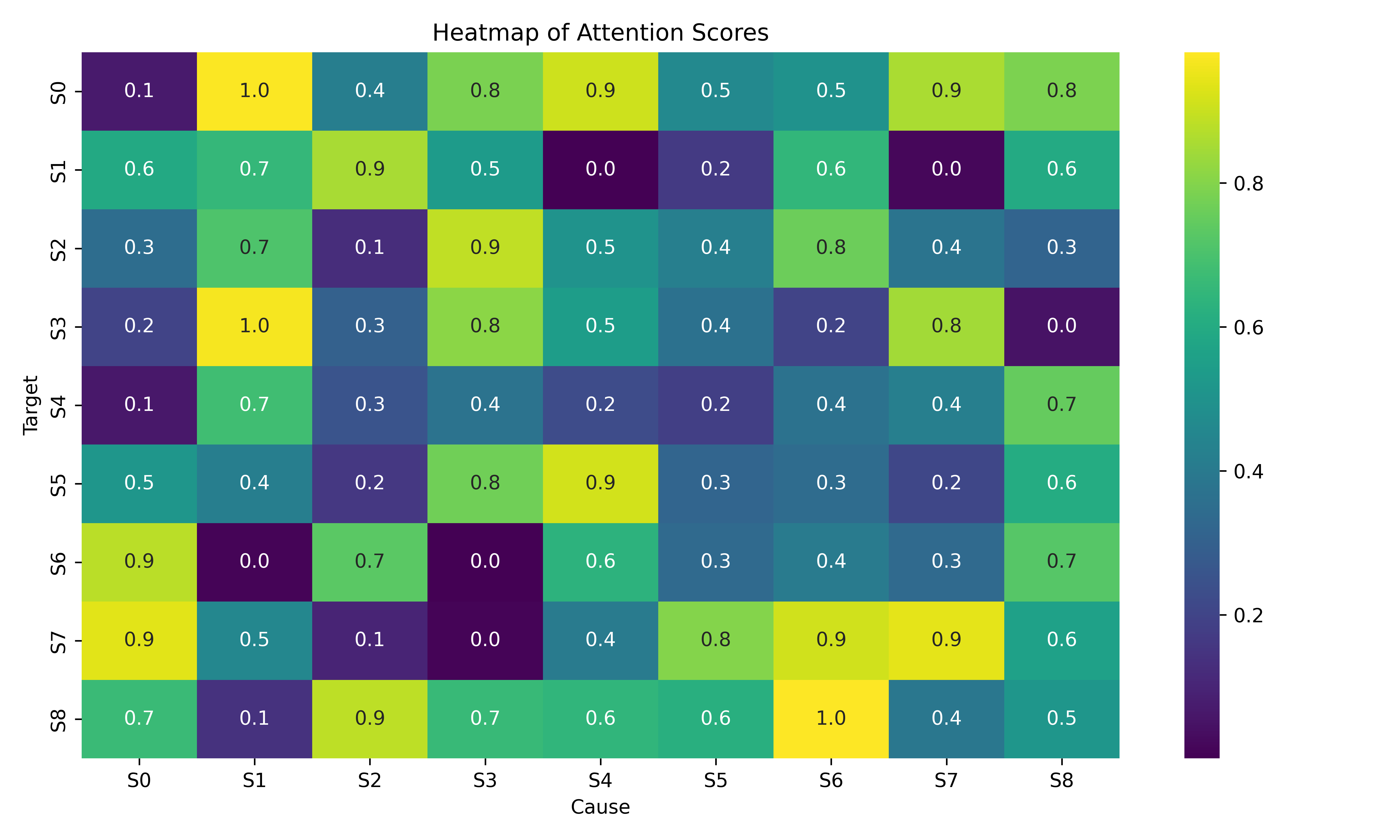}
    \caption{fMRI-9 variables. }
    \label{fig:fmri_heatmap}
\end{subfigure}
\caption{Attention score heat-maps showing variable relationships ((a): Finance dataset, (b): fMRI dataset)}
\label{fig:attention_heatmaps}
\end{figure}

\subsection{Comparative Insights}

The comparative analysis between the Finance and fMRI datasets reveals distinct characteristics in causal structure evolution. The Finance dataset exhibits a more heterogeneous distribution of causal relations, where some variables exert stronger influence over others. In contrast, the fMRI dataset shows more uniform connectivity, which aligns with the interconnected nature of neural activity.
Overall, the results confirm the effectiveness of our causal discovery model in uncovering meaningful relationships in both financial and neurological time series data. The progressive enhancement of the causal graphs demonstrates that longer training durations enable a more refined understanding of causal dependencies. This finding highlights the model’s capability to generalize across different domains, making it a robust tool for multivariate time series causal analysis.

\section{Evaluation Metrics for Causal Discovery:Comparative Study }
In this section, we present the evaluation metrics used to compare causal discovery models in multivariate time series (MTS).
Evaluating the performance of causal discovery models requires multiple metrics that assess various aspects of accuracy, reliability, and structural correctness. Below is a detailed comparison of key evaluation metrics used for causal discovery in MTS :F1-Score, Recall, Precision and Delay Estimation Accuracy (DEA) where:

\begin{itemize}

    \item \textbf{Delay Estimation Accuracy (DEA):}  
    This metric evaluates how accurately the model predicts the causal delays between time series. A higher accuracy indicates a better estimation of temporal dependencies.

    \begin{equation}
        \text{DEA} = 1 - \frac{1}{n} \sum_{i=1}^{n} |\hat{d}_i - d_i|
    \end{equation}
  Where \(d_i\) denotes the ground-truth delay of edge \(i\) and \(\hat d_i\) the delay predicted by DyCAST-Net, \(n\) is the number of true causal edges. 
\end{itemize}

Table 3 presents a comparative analysis of different causal discovery models applied to the Finance and fMRI datasets. The models are evaluated based on key performance metrics: F1 Score, Recall, and DEA. The results highlight the strengths and weaknesses of each approach in detecting causal relations and estimating delays.
\begin{table}[h]
    \centering
    \caption{Comparison of Model Performance on Finance and fMRI Datasets}
    \label{tab:model_performance}
    \begin{tabular}{|c|c|c|c|c|c|}
        \hline
        \textbf{Model} & \textbf{F1 (Finance)} & \textbf{F1 (FMRI)} & \textbf{Recall (Finance)} & \textbf{Recall (FMRI)} & \textbf{DEA} \\ 
        \hline
        \textbf{(DyCAST-Net} & \textbf{0.91} & \textbf{0.89} & \textbf{0.90} & \textbf{0.89} &  \textbf{0.89} \\ 
        \hline
        \textbf{TCDF} & 0.84 & 0.84 & 0.83 & 0.83 &  0.80 \\ 
        \hline
        \textbf{GCFormer} & 0.86 & 0.85 & 0.83 & 0.84 & 0.81 \\ 
        \hline
        \textbf{Causalformer} & 0.87 & 0.87 & 0.86 & 0.88 &  0.85 \\ 
        \hline
    \end{tabular}
\end{table}

\subsection{Analysis of Model Performance}
DyCAST-Net outperforms all other models, achieving the highest F1 Score (0.91 on Finance, 0.89 on fMRI) and Recall (0.90 on Finance, 0.89 on fMRI). These results confirm its superior ability to detect causal relationships. 
TCDF exhibits the weakest performance, with lower Recall and F1 Scores, indicating challenges in capturing causal links and leading to more false negatives. In contrast, CausalFormer provides a balanced performance (F1 = 0.87), demonstrating robustness in causal inference.
Delay estimation accuracy (DEA) is crucial for causal discovery. DyCAST-Net achieves the highest DEA (0.89), reinforcing its ability to model temporal dependencies accurately. CausalFormer follows with 0.85, while TCDF performs the worst (0.80), highlighting its limitations in delay estimation.
Overall, DyCAST-Net provides the best trade-off between F1 Score, Recall, and DEA, confirming its effectiveness in causal inference. CausalFormer remains a strong alternative, particularly in recall. GCFormer achieves moderate performance, outperforming TCDF but falling short of DyCAST-Net. 
The findings highlight DyCAST-Net as the most robust model for causal discovery in multivariate time series, balancing high accuracy and reliable delay estimation.

\section*{Conclusion}
We introduced DyCAST-Net, a deep learning framework unifying dilated convolutions and multi-head (sparse) attention for causal discovery and time series forecasting in complex, noisy data. Compared to methods like TCDF, MGCFORMER, and CausalFormer, it demonstrates robust performance and interpretable causal insights across both synthetic and real-world domains (e.g., finance, fMRI). By combining local feature extraction with global attention, DyCAST-Net achieves superior metrics (precision, recall, F1) and lower MSE while revealing fine-grained causal links. Additionally, it shows resilience to non-stationary signals, thanks to dynamic sparse attention and a flexible model design. Future work will focus on improving computational efficiency, incorporating transformer-inspired or hierarchical components, and adapting to irregular sampling and real-time learning. These enhancements will reinforce DyCAST-Net’s interpretability and scalability, paving the way for broader application in fields requiring reliable causal inference from dynamic systems.


\end{document}